\documentclass[conference,letterpaper]{IEEEtran}

\usepackage{amssymb}
\usepackage{graphicx}
\usepackage{mathrsfs}
\usepackage{url}
\usepackage{amsmath}
\usepackage{color}  
\usepackage[ruled]{algorithm2e}
\usepackage{listings}
\begin{document}

\title{An agent-driven semantical identifier using radial basis neural networks and reinforcement learning}

\author{\IEEEauthorblockN{Christian
Napoli\IEEEauthorrefmark{1,*},
Giuseppe Pappalardo \IEEEauthorrefmark{1}, and
Emiliano Tramontana\IEEEauthorrefmark{1}}
\IEEEauthorblockA{\IEEEauthorrefmark{1}Dpt. of Mathematics and
Computer Science, University of Catania, Italy}% <-this % stops
an unwanted space
\thanks{*Email: napoli@dmi.unict.it.}
\thanks{Proceedings of the XV Workshop "Dagli Oggetti agli Agenti"}}

% The paper headers
\markboth{An agent-driven semantical identifier using radial basis neural networks and reinforcement learning -- PREPRINT}%
{Shell \MakeLowercase{\textit{et al.}}: Bare Demo of
IEEEtran.cls for Journals}

 \begin{titlepage}
 \begin{center}
 {\Large \sc PREPRINT VERSION\\}
  \vspace{5mm}
{\huge An agent-driven semantical identifier using radial basis neural networks and reinforcement learning\\}
 \vspace{10mm}
 {\Large C. Napoli, G. Pappalardo, and E. Tramontana\\}
 \vspace{5mm}
{\Large \sc PUBLISHED ON: \bf Proceedings of the XV Workshop "Dagli Oggetti agli Agenti"}
 \end{center}
 \vspace{5mm}
 {\Large \sc BIBITEX: \\}
 
@inproceedings\{Napoli2014Anagent\\
year=\{2014\},\\
issn=\{1613-0073\},\\
url=\{http://ceur-ws.org/Vol-1260/\},\\
booktitle=\{Proceedings of the XV Workshop "Dagli Oggetti agli Agenti"\},\\
title=\{An agent-driven semantical identifier using radial basis neural networks and reinforcement learning\}, \\
publisher=\{CEUR-WS\},\\
volume=\{1260\},\\
author=\{Napoli, Christian and Pappalardo, Giuseppe and Tramontana, Emiliano\},\\
\}
 \vspace{5mm}
 \begin{center}
Published version copyright \copyright~2014  \\ %IEEE
\vspace{5mm}
UPLOADED UNDER SELF-ARCHIVING POLICIES\\
%NO COPYRIGHT INFRINGEMENT INTENDED \\
 \end{center}
\end{titlepage}

\title{An agent-driven semantical identifier using radial basis neural networks and reinforcement learning}
\author{\IEEEauthorblockN{Christian Napoli, Giuseppe Pappalardo, and Emiliano Tramontana}
\IEEEauthorblockA{Department of Mathematics and Informatics\\
University of Catania, Viale A. Doria 6, 95125 Catania, Italy\\
\{napoli, pappalardo, tramontana\}@dmi.unict.it}}

\maketitle

\begin{abstract}
  Due to the huge availability of documents in digital form, and the
  deception possibility raise bound to the essence of digital
  documents and the way they are spread, the authorship attribution
  problem has constantly increased its relevance.
  Nowadays, authorship attribution,
for both
  information retrieval and analysis, has gained great importance in
  the context of security, trust and copyright preservation.

  This work proposes an innovative 
multi-agent driven machine learning
  technique that has been developed for authorship attribution.
  By means of a preprocessing for word-grouping and time-period
  related analysis of the common lexicon, we determine a bias
  reference level for the recurrence frequency of the words within
  analysed texts, and then train a Radial Basis Neural Networks
  (RBPNN)-based classifier to identify the correct author.

  The main advantage of the proposed approach lies in the generality
  of the semantic analysis, which can be applied to different
  contexts and lexical domains, without requiring any modification.
  Moreover, the proposed system is able to incorporate an external
  input, meant to tune the classifier, and then self-adjust by means
  of continuous learning reinforcement.

\end{abstract}

\IEEEpeerreviewmaketitle

\section{Introduction}
Nowadays, the automatic attribution of a text to an author,
assisting both information retrieval and analysis,
has become an important issue, e.g.\ in the context of security, trust
and copyright preservation.  
This results from the availability
of documents in digital form, and the raising deception
possibilities bound to the essence of the digital reproducible
contents, as well as the need for new mechanical methods that can organise
the
constantly increasing amount of digital texts. 

During the last decade only, the field of text classification and
attribution has undergone new developement due to the novel
availability of computational intelligence techniques,  such as
natural language processing, advanced data mining and information
retrieval systems, machine learning and artificial intelligence
techniques, agent oriented programming, etc. 
Among such techniques,  Computer Intelligence (CI) and Evolutionary
Computation (EC) methods have been largely used for optimisation and
positioning problems~\cite{NapoliSCCI,Wozniak2012}.
In~\cite{bonanno2012optimal}, agent driven clustering
has been used as an advanced solution for some optimal management
problems, whereas in~\cite{Wozniak2008_3} such problems are solved for
mechatronical module controls.  Agent driven artificial intelligence
is often used in combination with advanced data analysis techniques in
order to create intelligent control systems~\cite{napoli2013hybrid,bonanno2014novel} by
means of multi resolution analysis~\cite{Wozniak2008_2}.  CI and
parallel analysis systems have been proposed in order to support
developers, as in~\cite{Napoli2013a,PappalardoT13,Tramontana13,NapoliWETICE}, where
such a classification and 
analysis was applied to assist refactoring in large software
systems~\cite{rosario2011, CalvagnaT13deliv, GiuntaPT12, Tramontana14}.

Moreover, CI and techniques like neural networks
(NNs) have been used in~\cite{capizzi2012innovative,bonanno2014cascade} in
order to model electrical networks and the related controls starting
by classification strategies, as well as for other complex physical
systems by using several kinds of hybrid NN-based approaches
\cite{napoli2010exploiting,capizzi2011hybrid,napoli2010hybrid,CapizziSP10}.  All
the said works use different forms of agent-based  modeling and
clustering for recognition purposes, and these methods efficiently
perform very challenging tasks, where other common computational
methods failed or had low efficiency or, simply, resulted as
inapplicable due to complicated model underlying the case study.  
In general, agent-driven machine learning has been proven as a
promising field of research for the purpose of text classification,
since it allows building classification rules by  means of
automatic learning, taking as a basis a set of known texts and
trying to generalise for unknown ones.

While machine learning and NNs are a very promising
field,  the effectiveness of such approaches
often lies on the correct and precise preprocessing of data, i.e.\
the definition of semantic categories, affinities and rules used to
generate a set of numbers characterising a text sample, to be
successively given as input to a classifier.  Typical text
classification, e.g.\ by using NNs, takes advantage of topics recognition,
however results are seldom appropriate
when it comes to classify people belonging to the same social group or
who are involved in a similar business (e.g.\ the classification of:
texts from different scientists in the same field of research, the
politicians belonging to the same party, 
 texts authored by different people using the same
technical jargon).

In our approach we devise a solution for
extracting from the analysed texts some
characteristics that can express the style of a
specific author.  Obtaining this kind of information
abstraction is crucial in order to create a precise and correct
classification system.  On the other hand, while data abound in the
context of text analysis,
a robust classifier should rely on input sets that are compact enough
to be apt to the training process.
Therefore, some data have to reflect averaged evaluations
that concern some anthropological aspects such as the historical
period, or the ethnicity, etc.
This work satisfies the above conditions of extracting compact data
from texts since we use a preprocessing tool for word-grouping and
time-period related analysis of the common lexicon.  Such a tool
computes   a bias reference system for the recurrence frequency of the
word used in the analysed texts.  The main advantage of this choice
lies in the generality of the implemented semantical identifier, which
can be then applied to different contexts and lexical domains without  requiring
any modification. 
Moreover, in order to have continuous updates or complete renewals of
the reference data, a statically trained NN would not
suffice to the purpose of the work.  For these reasons, the developed
system
is able to self-correct by means of continuous learning
reinforcement.  The proposed architecture also diminishes the human
intervention over time thanks to its self-adaption properties.
Our solution comprises three main collaborating \emph{agents}: the first for
\emph{preprocessing}, i.e.\ to extract meaningful data from texts; the second for 
\emph{classification} by means of a proper Radial Basis NN (RBNN); and
finally, one for \emph{adapting} by means of a feedforward NN.

The rest of this paper is  as follows.
Section~\ref{s:lexicon} gives the details of the implemented
preprocessing agent based on lexicon analysis.
Section~\ref{s:rbpnnclassifier} describes the proposed classifier agent
based on RBNNs, our introduced modifications and the structure of the
reinforcement learning agent.  Section~\ref{s:exp} reports
on the performed experiments and the related results.  Finally,
Section~\ref{s:related} gives a background of the existing related
works, while Section~\ref{s:conc} draws our conclusions.

\begin{figure}[!t]
  \centering
  \includegraphics[width=.48\textwidth]{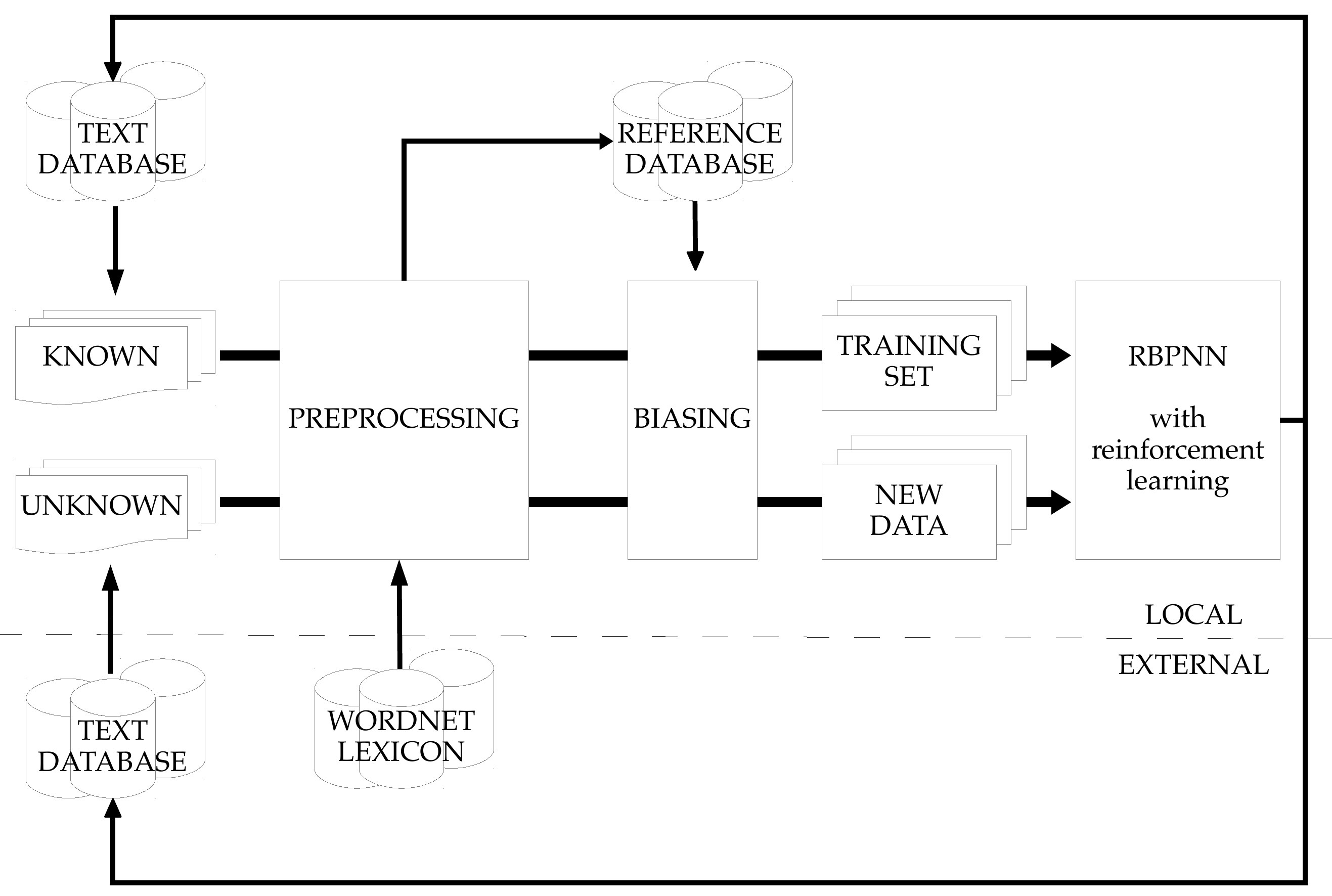}
  \caption{A general schema of the data flow through the agents of
    the developed system.} 
  \label{fig:inout}
\end{figure}

\section{Extracting semantics from lexicon}
\label{s:lexicon}

Figure~\ref{fig:inout} shows the agents for our developed system:
 a \emph{preprocessing agent}  extracts characteristics
from given text parts (see text database in the Figure), according to a
known set of words organised into groups (see reference database); 
a \emph{RBPNN agent} takes as input the extracted characteristics, properly
organised, and performs the identification on new data, after
appropriate training.  An additional agent, dubbed \emph{adaptive
  critic},  shown in Figure~\ref{fig:clearning},
dynamically adapts the behaviour of the \emph{RBPNN agent} when new
data are available.

Firstly, \emph{preprocessing agent}
analyses a text given as input
by counting the words that belong to a priori known groups of mutually
related words.  Such groups contain words that pertain to a given
concern, and have been built ad-hoc and
according to the semantic relations between words, hence e.g.\
assisted by the WordNet
lexicon\footnote{http://wordnet.princeton.edu}.

The fundamental steps of the said analysis (see also
Algorithm~\ref{algorithm1}) are the followings:
\begin{enumerate}
\item import a single text file containing the speech;
\item import word groups from a predefined database,
  the set containing all words from each group is called dictionary;
\item compare each word on the text with words on the dictionary;
\item if the word exists on the dictionary then the relevant
  group is returned;
\item if the word has not been found then search the available lexicon;
\item if the word exists on the lexicon then the related
  group is identified;
\item if the word is unkown, then a new lexicon is loaded and if the
  word is found then
   dictionary and  groups are updated;
\item search all the occurrences of the word in the text;
\item when an occurrence has been found, then remove it from the text
  and increase the group counter. 
\end{enumerate}

\begin{algorithm}[!t]\label{algorithm1}
\SetAlgoLined
\DontPrintSemicolon
Start,\;
Import  a speech into $Text$,\;
Load dictionary into $Words$,\;
Load group database into $Groups$,\;
$thisWord$ = Text.get();\;
\While{$thisWord$}
	{
		$thisGroup$ = Words.search($thisWord$);\;
		\If {!$thisGroup$}
			{ 
				Load a different $Lexicon$; \;
                                \If{$Lexicon$.exist($thisWord$)}
                                  {
                                  $Words$.update(); \;
                                  $Groups$.update(); \;
                                 }
                                 \Else
                                 {
                                   break;
                                 }}
			\While{\emph{Text.search($thisWord$)}}
			{
				Groups.count($thisGroup$);\;
			}
			$thisWord$ = Text.get();\;
	}
Export $Words$ and $Groups$, \;
Stop.\;
\caption{Find the group a word belongs to and count occurrences}
\end{algorithm}
\begin{figure}[!t]
  \centering
  \includegraphics[width=.42\textwidth]{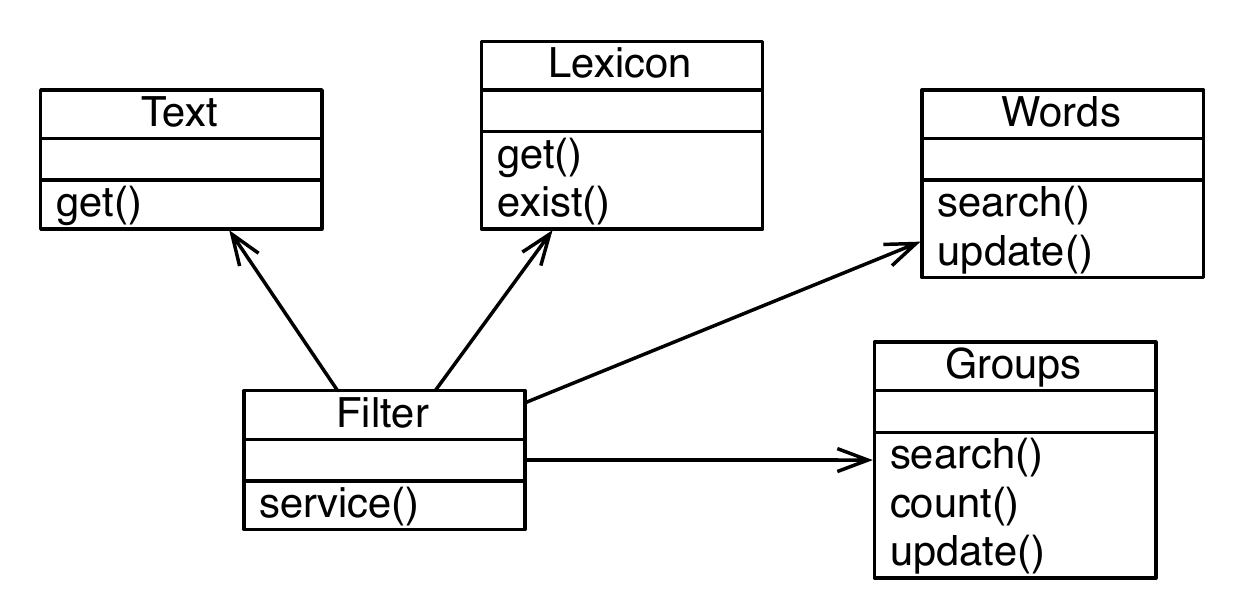}
  \caption{UML class diagram 
for handling groups and counting words belonging to a group.}
  \label{fig:uml}
\end{figure}
Figure~\ref{fig:uml} shows the UML class diagram for the software
system performing the above analysis.
Class \texttt{Text} holds a text to be analysed; class \texttt{Words} represents the
known dictionary, i.e.\ all the known words, which are organised into
groups given by class \texttt{Groups}; class  \texttt{Lexicon} holds 
several dictionaries.

\section{The RBPNN classifier agent}\label{s:rbpnnclassifier}
For the work proposed here, we use a variation on Radial Basis Neural Networks
(RBNN). RBNNs have a topology similar to common FeedForward Neural Networks
(FFNN) with BackPropagation Training Algorithms (BPTA): the primary
difference only lies in the activation function that, instead of being
a sigmoid function or a similar activation function, is a statistical
distribution or a statistically significant mathematical function.
The selection of transfer functions is indeed decisive for the speed of 
convergence in approximation and classification problems~\cite{duch07}.
The kinds of activation functions used for Probabilistic Neural
Networks (PNNs) have to meet some
important properties to preserve the generalisation abilities of the
ANNs.  In addition, these functions have to preserve the decision
boundaries of the probabilistic neural networks.  
The selected RBPNN architecture is shown in Figure~\ref{fig:rbpnn} and takes advantage
from both the PNN topology and the Radial Basis Neural Networks (RBNN)
used in~\cite{Bonanno12}.

Each neuron performs a weighted sum of its inputs and passes it
through a transfer function f to produce an output.  This occurs for each neural
layer in a FFNN.  The network can be perceived as a model connecting
 inputs and outputs, with the weights and thresholds being free
parameters of the model, which are modified by the
training algorithm.  Such networks can model functions of almost
arbitrary complexity with the number of layers and the number of units
in each layer determining the function complexity.  A FFNN is capable
to generalise the model, and to separate the input space in various
classes (e.g.\ in a 2D variable space it is equivalent to the
separation of the
different semi-planes).  In any case, such a FFNN can only create a
general model of the entire variable space, while can not insert single
set of inputs into categories.  
On the other hand, a RBNN is capable of clustering the inputs by
fitting each class by means of a radial basis function~\cite{haykin},
while the model is not general for the entire variable space, it is
capable to act on the single variables (e.g.\ in
a 2D variable space it
locates closed subspaces, without any inference on the remaining space
outside such
subspaces).

\begin{figure}[!t]
  \centering
  \includegraphics[width=.34\textwidth]{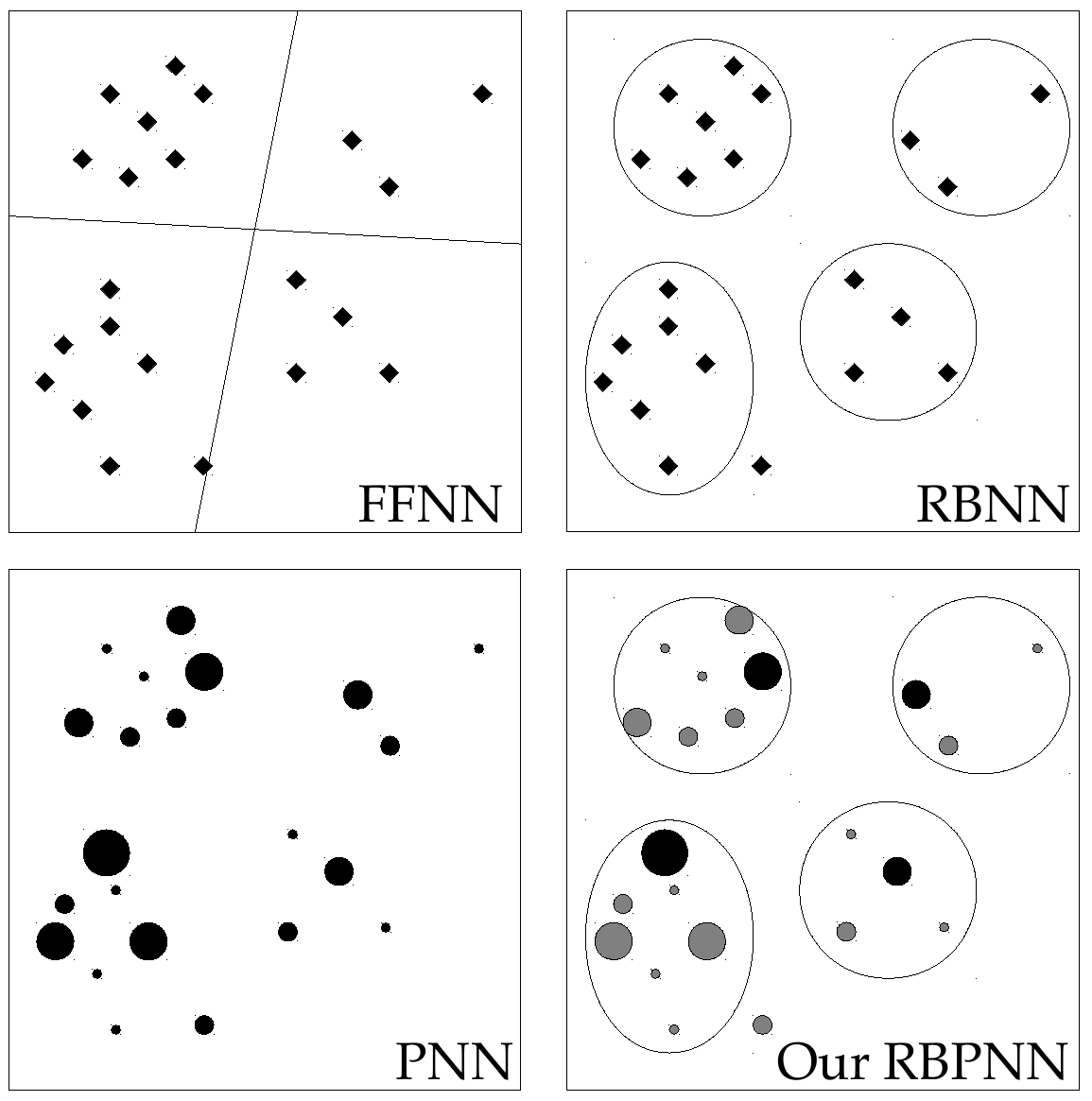}
  \caption{A comparison between results of several types of NNs, Our
    RBPNN includes the maximum probability selector module}
  \label{fig:OurRBPNN}
\end{figure}

Another interesting topology is provided by PNNs, which are mainly FFNNs also functioning as Bayesian
networks with Fisher Kernels~\cite{FisherK}.  By replacing the sigmoid
activation function often used in neural networks with an exponential
function, a PNN can compute nonlinear decision boundaries approaching
the Bayes optimal classification~\cite{PNNs}.  Moreover, a PNN generates
accurate predicted target probability scores with a probabilistic
meaning (e.g.\ in the 2D space it is equivalent to attribute a
probabilistic score to some chosen points, which in 
Figure~\ref{fig:OurRBPNN} are represented as the size of the
points). 

Finally, in the presented approach we decided to combine the
advantages of both RBNN and PNN using the so called RBPNN. The RBPNN
architecture, while preserving the capabilities of a PNN, due to its
topology, then being capable of statistical inference, is also
capable of clustering since the standard activation functions of a PNN
are substituted by radial basis functions still verifying the Fisher
kernel conditions required for a PNN (e.g.\ such an architecture in the
2D variable space can both locate subspace of points and give to
them a probabilistic score).
Figure~\ref{fig:OurRBPNN} shows a representation of the behaviour for
each network topology presented above.

\subsection{The RBPNN structure and topology}\label{s:rbpnnT}
In a RBPNN both the input and the first hidden layer exactly match the
PNN architecture: the input neurones are used as distribution units
that supply the same input values to all the neurones in the first
hidden layer that, for historical reasons, are called \emph{pattern
  units}.  In a PNN, each pattern unit performs the dot product of the
input pattern vector $\mathbf{v}$ by a weight vector
$\mathbf{W}^{(0)}$, and then performs a nonlinear operation on the
result.  This nonlinear operation gives output $\mathbf{x}^{(1)}$ that
is then provided to the following summation layer.  While a common
sigmoid function is used for a standard FFNN with BPTA, in
a PNN
the activation function is an exponential, such that, for the
$j$-esime neurone the output is
\begin{equation}
  \mathbf{x}^{(1)}_j\propto
  \exp\left(\frac{||\mathbf{W}^{(0)}\cdot\mathbf{v}||}{2\sigma^2}\right) 
\label{eq:exp}
\end{equation}
where $\sigma$ represents the statistical distribution spread.

The given activation function can be modified or substituted while the
condition of Parzen (window function) is still satisfied for the
estimator $\hat{N}$.  In order to satisfy such a condition some rules
must be verified for the chosen window function in order to obtain the
expected estimate, which can be expressed as a Parzen window estimate
$p(x)$ by means of the kernel $K$ of $f$ in the $d$-dimensional space
$S^d$ 
\begin{equation}
\begin{array}{l}
p_n(x)=\frac{1}{n} \sum\limits_{i=1}^n \frac{1}{h_n^d} K\left(\frac{x-x_i}{h_n}\right) \\ \\
\int\limits_{S^d} K(x) dx =1
\end{array}
\label{eq:window}
\end{equation}
where $h_n \in \mathbb{N}$ is called window width or bandwidth
parameter and corresponds to the width of the kernel. In general $h_n
\in \mathbb{N}$ depends on the number of available sample data $n$ for
the estimator $p_n(x)$. 
Since the estimator $p_n(x)$ converges in mean square to the expected value $p(x)$ if
\begin{equation}
\begin{array}{rcl}
\lim \limits_{n\rightarrow\infty} \langle p_n(x) \rangle &=& p(x)\\ \\
\lim \limits_{n\rightarrow\infty} \mbox{var}\left(p_n(x)\right) &=& 0
\end{array}
\label{eq:estimator}
\end{equation}
where $\langle p_n(x) \rangle$ represents the mean estimator values
and $\mbox{var}\left(p_n(x)\right)$ the variance of the estimated
output with respect to the expected values, the  Parzen condition
states that such convergence holds within the following conditions: 
\begin{equation}
\begin{array}{rcl}
\sup \limits_x K(x) &<& \infty\\ \\
\lim \limits_{|x|\rightarrow\infty} xK(x) &=& 0\\ \\
\lim \limits_{n\rightarrow\infty} h_d^n &=& 0\\ \\
\lim \limits_{n\rightarrow\infty} nh_d^n &=& \infty
\end{array}
\label{eq:parzen}
\end{equation}
In this case, while preserving the PNN topology, to obtain the RBPNN
capabilities, the activation function is substituted with a radial
basis function (RBF); an RBF still verifies all the conditions stated
before.  It then follows the equivalence between the
$\mathbf{W}^{(0)}$ vector of weights and the centroids vector of a
radial basis neural network, which, in this case, are computed as the
statistical centroids of all the input sets given to the network.
\begin{figure}[!t]
  \centering
  \includegraphics[width=.48\textwidth]{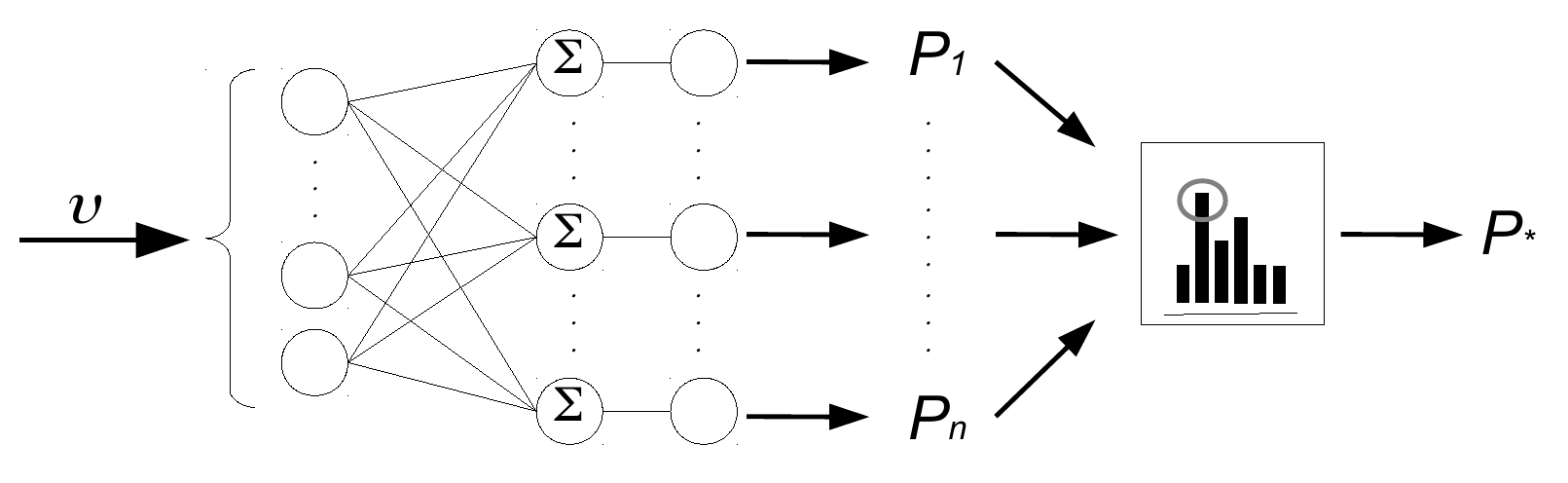}
  \caption{A representation of a Radial Basis Probabilistic Neural
    Network with maximum probability selector module}
  \label{fig:rbpnn}
\end{figure}

We name $f$ the chosen radial basis function, then the new output of
the first hidden layer for the j-esime neurone is
\begin{equation}
  \mathbf{x}^{(1)}_j\triangleq f \left(\frac{||\mathbf{v}-\mathbf{W}^{(0)}||}{\beta}\right)
  \label{eq:rbf} 
\end{equation}
where $\beta$ is a parameter that is intended to control the
distribution shape, quite similar to the $\sigma$ used in
(\ref{eq:exp}).

The second hidden layer in a RBPNN is identical to a PNN, it just
computes weighted sums of the received values from the preceding
neurones. This second hidden layer is called indeed summation layer:
the output of the k-esime summation unit is
\begin{equation}
  \mathbf{x}^{(2)}_k= \sum_j \mathbf{W}_{jk} \mathbf{x}^{(1)}_j
  \label{eq:sum}
\end{equation}
where $\mathbf{W}_{jk}$ represents the weight matrix.  Such weight
matrix consists of a weight value for each connection from the j-esime
pattern units to the k-esime summation unit.  These summation units
work as in the neurones of a linear perceptron network.  The training
for the output layer is performed as in a RBNN, however since the
number of summation units is very small and in general remarkably less
than in a RBNN, the training is simplified and the speed greatly
increased~\cite{Deshuang96}.

The output of the RBPNN (as shown in Figure~\ref{fig:rbpnn}) is given
to the maximum probability selector module, which effectively acts as
a one-neuron output layer. This selector receives as input the
probability score generated by the RBPNN and attributes to one author
only the analysed text, by selecting the most probable author, i.e.\
the one having the maximum input probability score.  Note that  the links
to this selector are weighted (with weights adjusted during the
training), hence the actual input  is the product between the weight
and the output of the summation layer of the RBPNN.

\subsection{Layer size for a RBPNN}
\begin{figure}[!t]
  \centering
  \includegraphics[width=.48\textwidth]{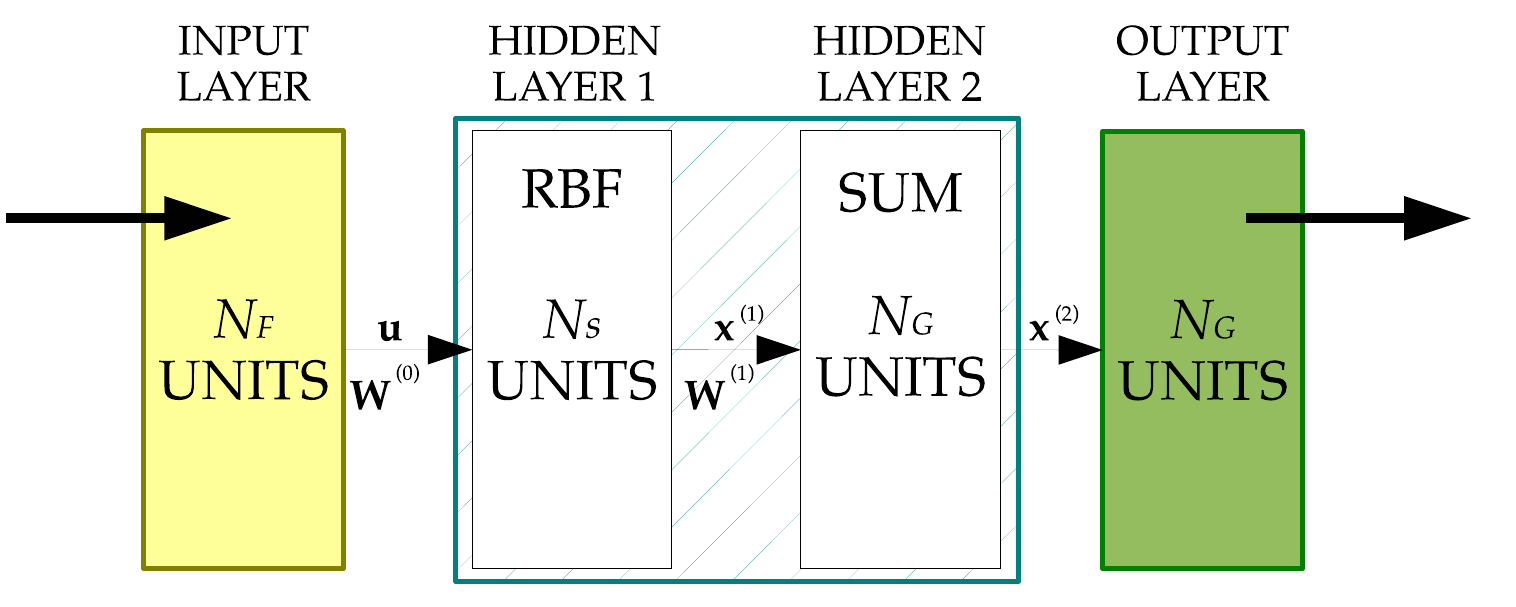}
  \caption{Setup values for the proposed RBPNN: $N_F$ is the number of
    considered lexical groups, $N_S$ the number of analysed texts, and
    $N_G$ is the number of people that can possibly be recognised as authors.}
  \label{fig:RBPNN2}
\end{figure}

The devised topology enables us to distribute to different layers
of the network  different parts of the classification task.  While the
pattern layer is just a nonlinear processing layer, the summation
layer selectively sums the output of the first hidden layer.  The
output layer fullfills the nonlinear mapping such as classification,
approximation and prediction.  In fact, the first hidden layer of the
RBPNN has  the responsibility to perform the fundamental task
 expected from a neural network~\cite{Zhao03}.
In order to have a proper classification of the input dataset, i.e.\
of analysed texts to be attributed to authors,
the size of the input layer should match the
exact number $N_F$ of 
different lexical groups
given to the RBPNN, whereas the size of
the pattern units should match the number of samples, i.e.\ analysed texts, $N_S$.
The number of the summation units in the second hidden layer is equal
to the number of output units, these should match the number of people
$N_G$ we are interested in for the correct recognition of the speakers
(Figure~\ref{fig:RBPNN2}).

\subsection{Reinforcement learning}
In order to continuously update the reference database for our
 system, a statically trained NN would not
suffice for the purpose of the work.  Since the aim of the presented
system is having an expanding database of text samples for
classification and recognition purpose, the agent driven
identification should dynamically follow the changes in such a
database.  When a new entry is made then the related feature set and
biases change, it implies that also the RBPNN should be properly
managed in order to ensure a continuous adaptive control for
reinforcement learning.
Moreover, for the considered domain it is desirable that a human
supervisor supply suggestions, expecially when the system starts
working.
The human activities are related to the supply of new entries into the
text sample database, and to the removal of misclassifications made by
the RBPNN.

\begin{figure}[t]
  \centering
  \includegraphics[width=.48\textwidth]{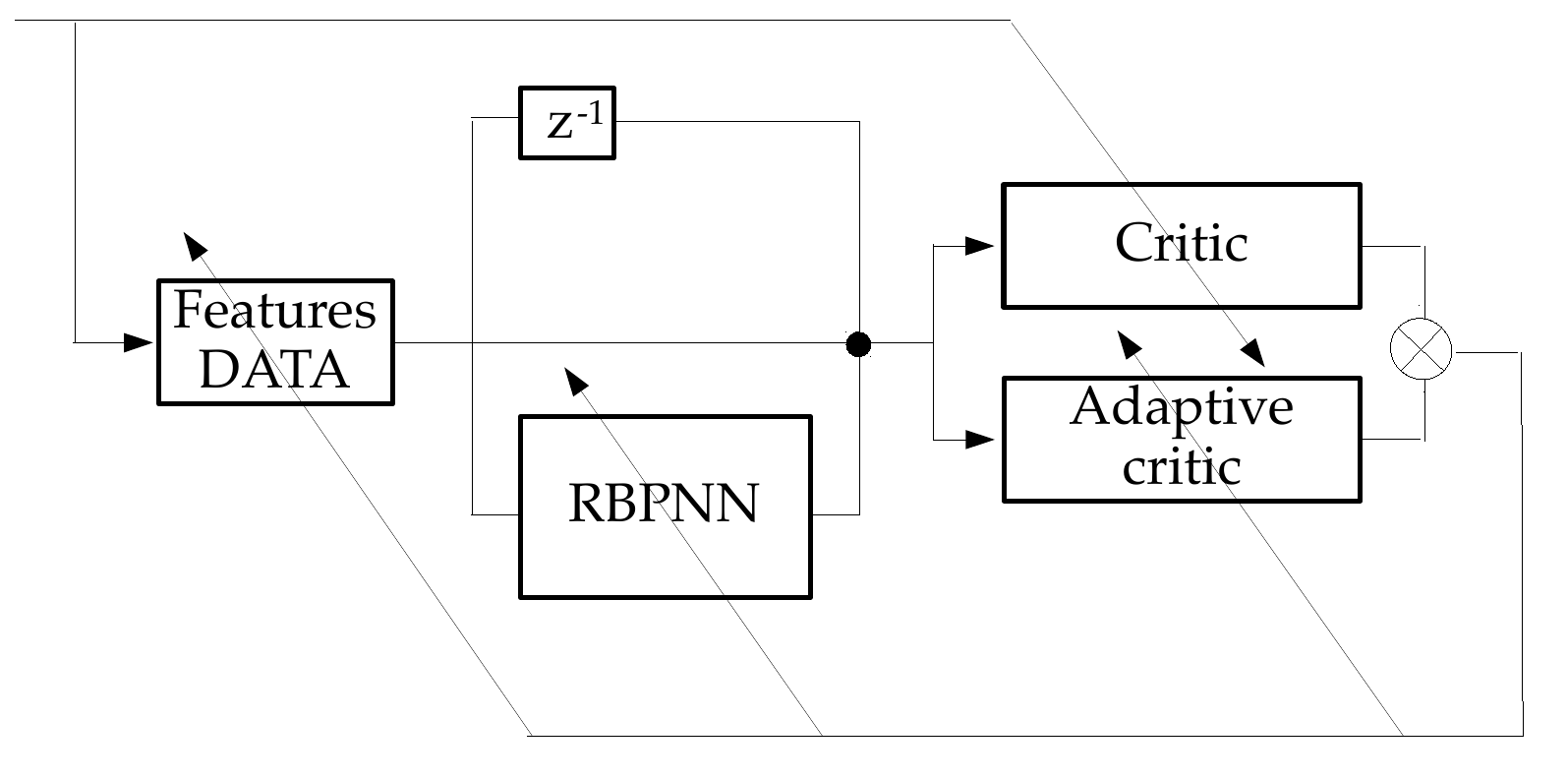}
  \caption{The adopted supervised  learning model
    reinforcement. Slanting arrows represent internal commands
    supplied in order to control  or change the status of the modules,
    straight arrows represent the data flow along the model,  $z^{-1}$
    represents  a time delay module which provides 1-step delayed
    outputs.} 
  \label{fig:clearning}
\end{figure}

We used a supervised control configuration (see
Figure~\ref{fig:clearning}), where the external control is provided by
the actions and choices of a human operator.  While the RBPNN is
trained with a classical backpropagation learning algorithm, it  is
also embedded into an
actor-critic reinforcement learning architecture, which back propagates
learning by evaluating the correctness of the RBPNN-made choices with
respect to the real word.

Let $\xi$ be the error function, i.e.\ $\xi=0$ for the results supported
by human verification, or the vectorial deviance for the results
not supported by a positive human response. 
This assessment is made by an agent named \emph{Critic}.
We consider 
 the filtering step for
 the RBPNN output, to be both:  \emph{Critic}, i.e.\ a human supervisor 
acknowledging or rejecting RBPNN  classifications; or  
\emph{Adaptive critic}, i.e.\ an agent embedding a NN that in the long run simulates the
control activity made by the human \emph{Critic}, hence decreasing  human
 control over time.   \emph{Adaptive critic} needs to
learn, and this learning is obtained by a modified backpropagation
algorithm  using just $\xi$ as error function.  
Hence,
\emph{Adaptive critic} has been implemented by a simple feedforward
NN trained, by means of a traditional gradient descent
algorithm so that the weight modification $\Delta w_{ij}$ is 
\begin{equation}
\Delta w_{ij} = -\mu \frac{\partial \xi}{\partial w_{ij}} = -\mu
\frac{\partial \xi}{\partial \tilde{f}_{i}} \frac{\partial
  \tilde{f}_{i}}{\partial \tilde{u}_{i}} \frac{\partial
  \tilde{u}_{i}}{\partial w_{ij}} 
\end{equation}
The $\tilde{f}_i$ is the activation of $i$-esime neuron, $\tilde{u}_i$
is the $i$-esime input to the neurone weighted as 
\begin{equation}
\tilde{u}_i = \sum_j w_{ij} \tilde{f}_j(\xi_i)
\end{equation}

The result of the adaptive control determines whether to continue the
training of the RBPNN with new data, and whether the
last training results should be saved or discarded.  At runtime this
process results in a continuous adaptive learning, hence avoiding the
classical problem of NN polarisation and overfitting. 
Figure~\ref{fig:clearning} shows
the developed learning system reinforcement. According to the
literature~\cite{cl1,cl2,cl3,cl4},  straight lines represent the
data flow, i.e.\  training data fed to the RBPNN, then new data inserted by
a supervisor, and the output of the RBPNN sent to the Critic modules
also by means of a delay operator $z^{-1}$. Functional modifications
operated within the system are represented as slanting arrows, i.e.\  the
choices made by a human supervisor (\emph{Critic})
modify the \emph{Adaptive critic}, which adjust the weight of its
NN; the combined output of \emph{Critic} and
\emph{Adaptive critic} determines whether the RBPNN should undergo
more training epochs and so modify its weights.

\section{Experimental setup}\label{s:exp}

\begin{figure*}[!t]
  \centering
  \includegraphics[width=.48\textwidth]{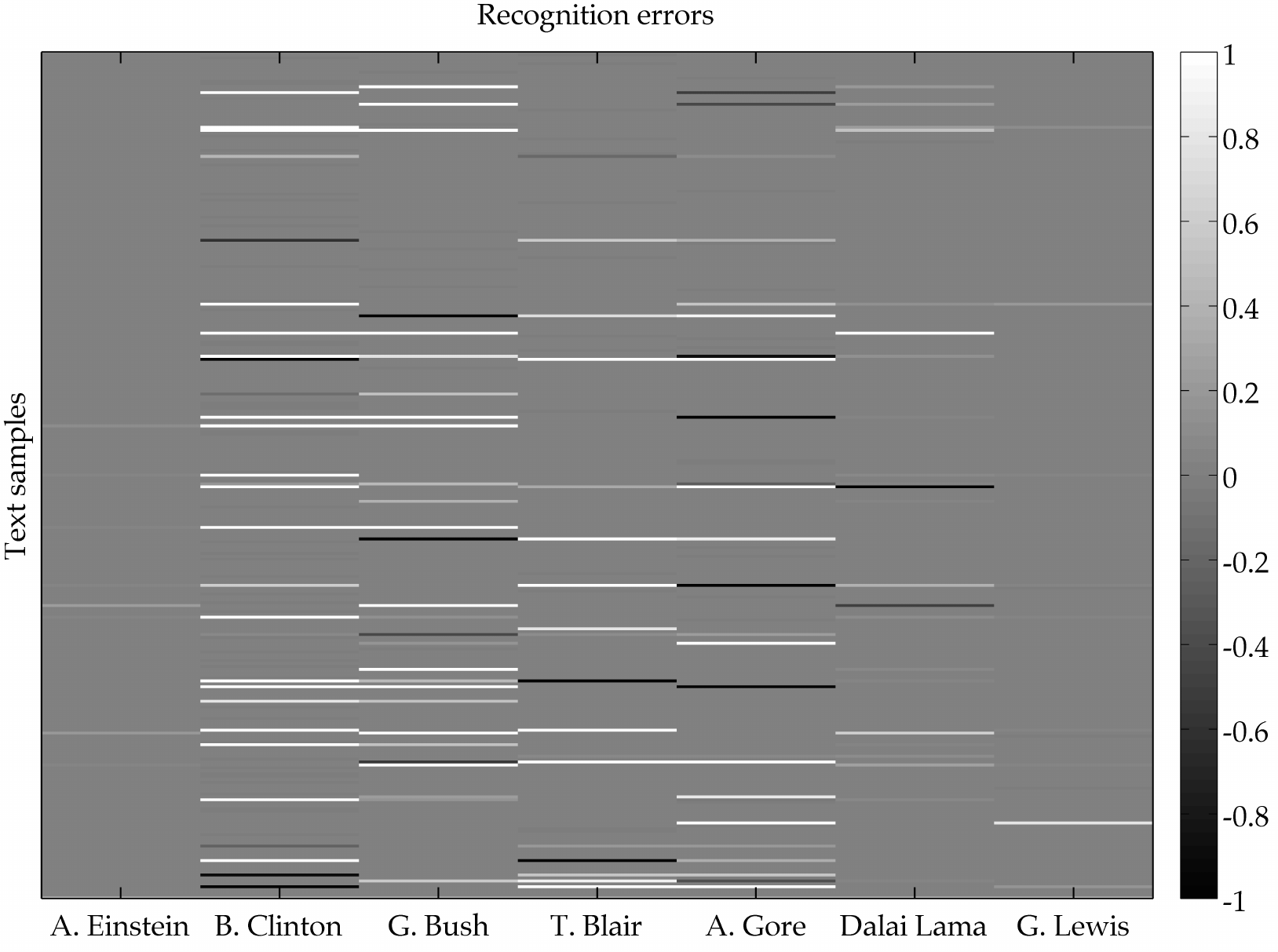}~~
    \includegraphics[width=.48\textwidth]{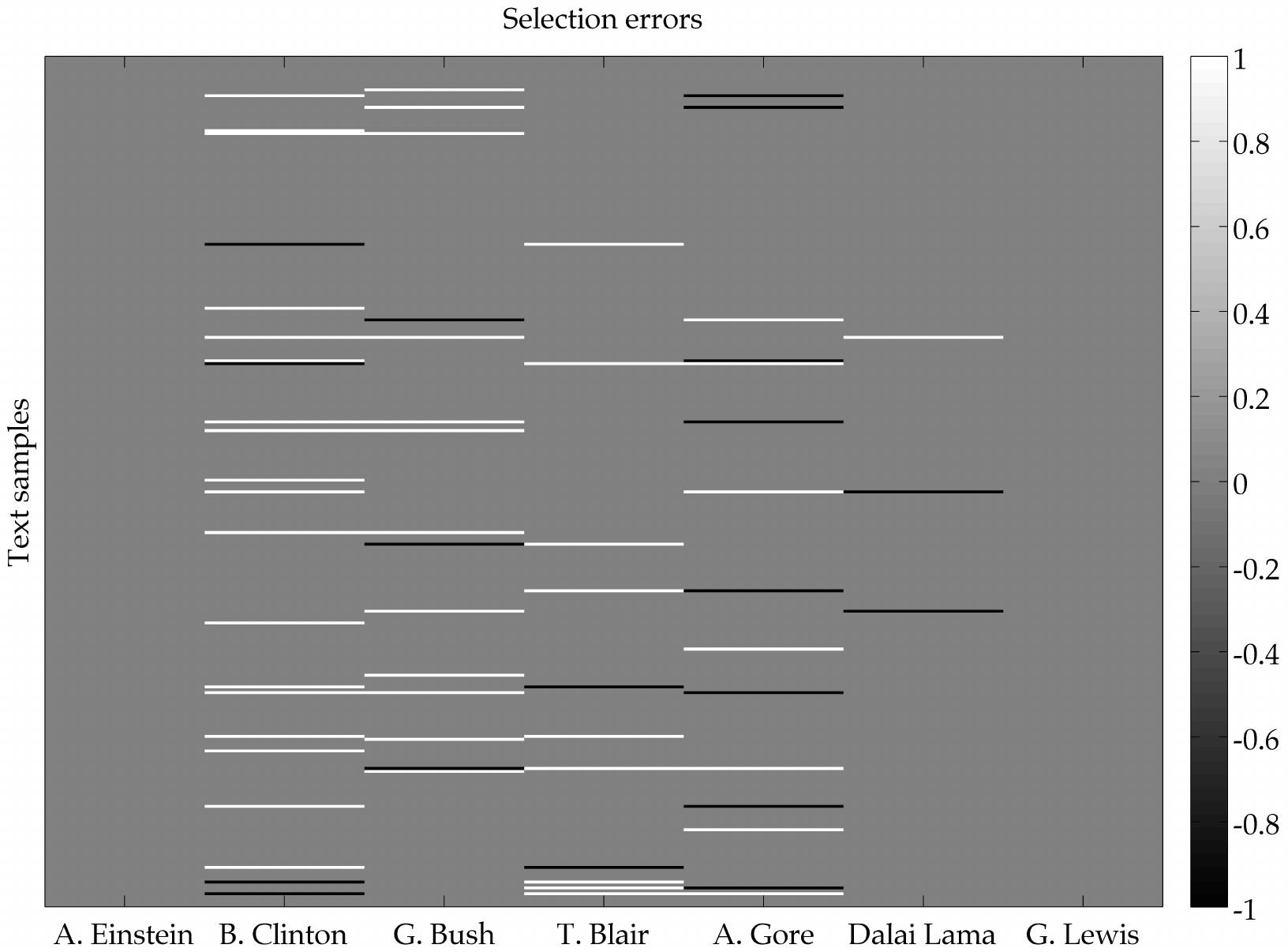}
  \caption{The obtained performance for our classification
    system before (left) and after (right) the maximum probability
    selector choice. The mean grey color represents the correct
    classifications, while white color represents missed
    classification and black color false classifications.} 
  \label{fig:sf}
\end{figure*}

The proposed RBPNN architecture has been tested using several text
samples collected from public speeches of different people both from
the present and the past era.  Each text sample has been given to the
\emph{preprocessing agent}
that extract some characteristics  (see Section~\ref{s:lexicon}), then
such results 
 have been given to the \emph{classification agent}.
The total number of text samples  was 344, and we used 258 of them for
training the  \emph{classification agent} and 86 for validation.  The
text samples, both for training
and validation, were from different persons that have given a speech
(from A.\ Einstein to G. Lewis), as shown in
Figure~\ref{fig:sf}.

Given the flexible structure of the implemented learning model, the
word groups are not fixed and can be modified, added or removed over
time by an external tuning activity. 
By using the count of words in a group,
 instead of a word-by-word counts,
 the multi-agent system realises a statistically driven classifier that
identifies the main semantic concerns regarding the text samples, and
then, attributes such concerns to the most probable person.

The relevant information useful  in order to recognise the author of
the speech is usually  largely spread over a certain number of word groups
 that could be indication of the cultural extraction,
heritage, field of study, professional category etc.  This implies
that we can not exclude any word group, a priori, while the RBPNN
could learn to automatically enhance the relevant information in order
to classify the speeches. 

Figure~\ref{fig:sf}-left shows an example of the classifier
performances for results generated by the RBPNN (before the filter
implemened by the probabilistic selector).
Since the RBPNN results have a probability between 0 and 1, then the
shown performance is 0 when a text was correctly attributed (or not
attributed) to a specific person.
Figure~\ref{fig:sf}-right shows the performances of the system when
including the probabilistic selector.
For this case, a boolean selection is involved, then correct
identifications are represented as 0, false positive identifications
as $-1$ (black marks), and missed identifications as $+1$ (white
marks).

For validation purposes,
Figure~\ref{fig:sf}-(left and right) shows results according to $e$:
\begin{equation}
e=y-\tilde{y}
\end{equation}
where $e$ identifies the performance, $\tilde{y}$ the classification
result, and $y$ the expected result.  
Lower $e$ (negative values) identify an excess of confidence in the
attribution of a text to a person, while greater $e$ (positive values)
identify a lack of confidence in that sense.

The system was able to correctly attribute the text to the proper
author with only a 20\% of missing assignments.

\section{Related Works}\label{s:related}

Several generative models can be used to characterise datasets that
determine properties and allow grouping  data into classes.
Generative models are based on stochastic
block structures~\cite{Nowicki01}, or on `Infinite Hidden Relational
Models'~\cite{Zhao06}, and `Mixed Membership Stochastic
Blockmodel'~\cite{Airoldi09}.  The main issue of class-based models is
the type of relational structure that such solutions are capable to
describe.  Since the definition of a class is attribute-dependent,
generally the reported models risk to replicate the existing classes
for each new attribute added.
E.g.\ such models would be unable to efficiently organise similarities
between the classes `cats' and `dogs' as child classes of the more
general class `mammals'.  Such attribute-dependent classes would have
to be replicated as the classification generates two different classes
of `mammals': the class `mammals as cats' and the class `mammals as
dogs'.  Consequently, in order to distinguish between the different
races of cats and dogs, it would be necessary to further multiply the
`mammals' class for each one of the identified race.  
Therefore, such models quickly lead to an explosion of classes.  In
addition, we would either have to add another class to handle each
specific use or a mixed membership model, as for crossbred species.

Another paradigm concerns the 'Non-Parametric Latent Feature Relational
Model'~\cite{Miller09}, i.e.\ a Bayesian nonparametric model in which each
entity has boolean valued latent features that influence the model's
relations.  Such relations depend on well-known covariant sets, which
are neither explicit or known in our case study at the
moment of the initial analysis.

In~\cite{novo07}, the authors propose a sequential forward feature
selection method to find the subset of features that are relevant to a
classification task.  This approach uses novel estimation of the
conditional mutual information between candidate feature and classes,
given a subset of already selected features used as a
classifier independent criterion for evaluating  feature subsets.  

In~\cite{bonanno2012some}, data from the charge-discharge simulation
of lithium-ions battery energy storage are used for classification
purposes with recurrent NNs and PNNs by means of a theoretical
framework based on signal theory.

While showing the effectiveness of the neural network based
approaches, in our case study classification results are given by
means of a probability, hence the use of a RBPNN, and an on-line
training achieved by reinforcement learning.

\section{Conclusion}\label{s:conc}

This work has presented a multi-agent system, in which an agent
analyses fragments of texts and another agent consisting of a RBPNN
classifier, performs probabilistic clustering.  The system has
successfully managed to identify the most probable author among a
given list for the examined text samples.
The provided identification can be used in order to complement and
integrate a comprehensive verification system, or other kinds of
software systems  trying to automatically identify the author of a
written text.  The RBPNN classifier agent is continuously trained by means of
reinforcement learning techniques in order to follow a potential
correction provided by an human supervisor, or an agent that learns
about supervision.
The developed system was
also able to cope with new data that are continuously fed into the
database, for the adaptation abilities of its collaborating agents and
their reasoning based on NNs.

\section*{Acknowledgment}

This work has been supported by project PRIME funded within POR FESR
Sicilia 2007-2013 framework and project PRISMA PON04a2 A/F funded by
the Italian Ministry of University and Research within PON 2007-2013
framework.

\bibliography{napoli}{}
\bibliographystyle{IEEEtran}
\end{document}